\documentclass{article}
\usepackage{graphics} 
\usepackage{epsfig} 
\usepackage{multirow}
\usepackage{booktabs} %
\usepackage{array}
\usepackage{amsmath} 
\usepackage{amssymb}  
\usepackage{lipsum}
\usepackage{pifont}
\usepackage{graphicx}
\usepackage{subcaption}
\usepackage{wrapfig}

\usepackage[final]{corl_2025} 

\title{Ensuring Force Safety in Vision-Guided Robotic Manipulation via Implicit Tactile Calibration}

\author{
  Lai Wei$^{*}$ \\
  Sun Yat-sen University; UC San Diego \\
  \texttt{law016@ucsd.edu} \\
  \And
  Jiahua Ma$^{*}$ \\
  Sun Yat-sen University \\
  \texttt{majiahua99@gmail.com} \\
  \And
  Yibo Hu \\
  Zhejiang University \\
  \texttt{boyihu@zju.edu.cn  } \\
  \And
  Ruimao Zhang$^{\dagger}$ \\
  Sun Yat-sen University\\
  \texttt{zhangrm27@mail.sysu.edu.cn} \\
}

\begin{document}
\maketitle

\renewcommand*{\thefootnote}{}
\footnotetext[1]{$^*$Equal contribution. $^\dagger$Corresponding author. }

\vspace{-7mm}
\begin{abstract}
In unstructured environments, robotic manipulation tasks involving objects with constrained motion trajectories—such as door opening—often experience discrepancies between the robot's vision-guided end-effector trajectory and the object's constrained motion path. 
Such discrepancies generate unintended harmful forces, which, if exacerbated, may lead to task failure and potential damage to the manipulated objects or the robot itself. 
To address this issue, this paper introduces a novel diffusion framework, termed SafeDiff. 
Unlike conventional methods that sequentially fuse visual and tactile data to predict future robot states, our approach generates a prospective state sequence based on the current robot state and visual context observations, using real-time force feedback as a calibration signal. 
This implicitly adjusts the robot’s state within the state space, enhancing operational success rates and significantly reducing harmful forces during manipulation. 
Additionally, we develop a large-scale simulation dataset named SafeDoorManip50k, offering extensive multimodal data to train and evaluate the proposed method. 
Extensive experiments show that our visual-tactile model substantially mitigates the risk of harmful forces in the door opening task, across both simulated and real-world settings. Project page is available at \href{https://i-am-future.github.io/safediff/}{this URL}.

\end{abstract}



\section{Introduction}
\label{sec:introduction}

\begin{wrapfigure}{r}{0.50\textwidth} 
\vspace{-10pt} 
\centering
\includegraphics[width=0.48\textwidth]{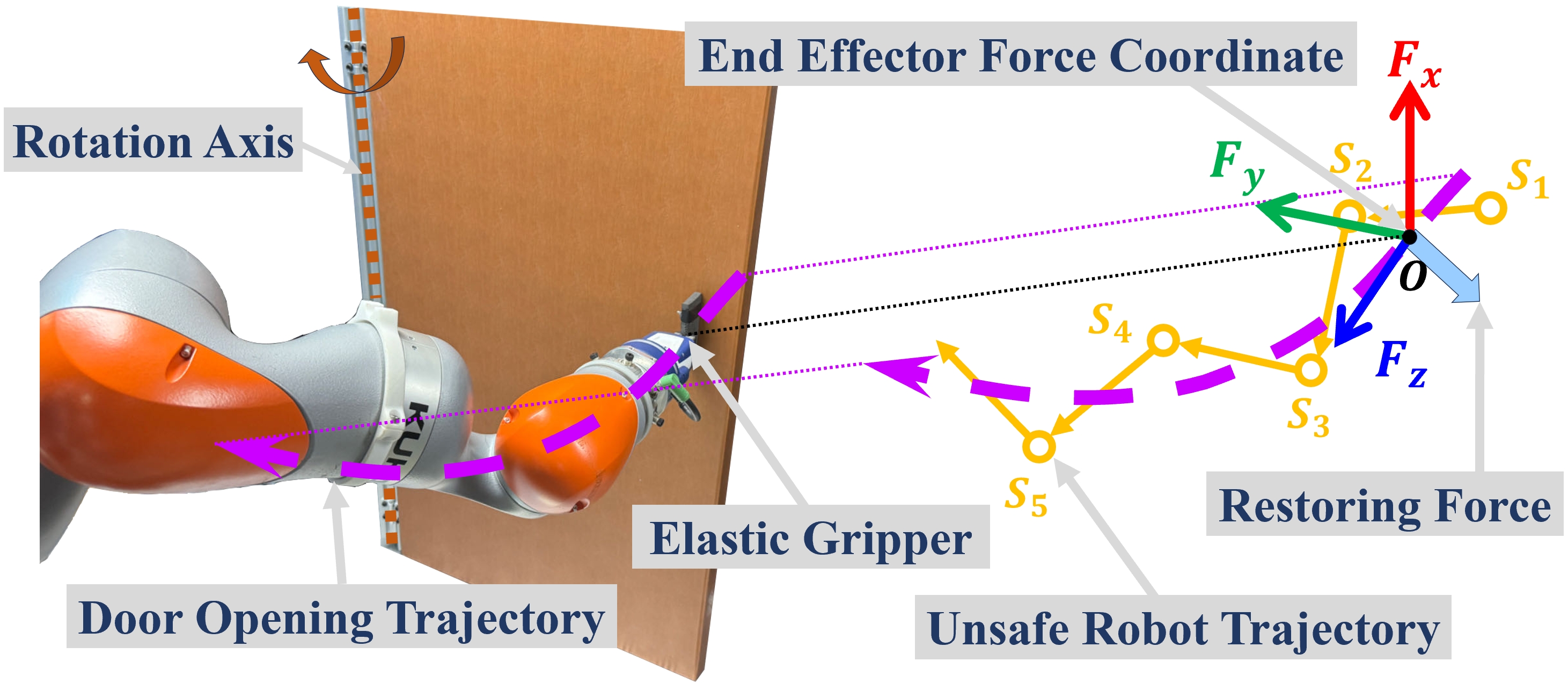} 
\caption{The restoring force exerted by the robot’s end-effector can be decomposed into three components: $F_x$, $F_y$, and $F_z$. The component $F_z$ is tangent with the door’s opening trajectory and is termed the \textbf{effective force}. The forces lying in the xOy plane are orthogonal to the trajectory. These forces might cause damage to both the robot and the door and are referred to as \textbf{harmful forces}.} 
\label{fig:intro}
\vspace{-10pt} 
\end{wrapfigure}
In industrial and everyday settings, robotic manipulation tasks often involve objects whose motion trajectories are inherently constrained, such as opening doors, closing windows, pulling drawers, or assembling bolts and pins. Under visual guidance, the motion trajectory generated for the robot end-effector may deviate from the constrained trajectory of the manipulated object, leading to unintended additional forces at the end-effector, as illustrated in Fig. \ref{fig:intro}. As these mismatches increase, they can cause task failure, and the resulting harmful forces may even damage the manipulated object or the robot’s joint motors. Therefore, ensuring precise force regulation is critical for both safe and efficient manipulation. In this work, we define that maintaining the harmful forces during manipulation remain within a safe threshold as \textbf{force safety}.
  
Traditional approaches to force safety primarily rely on impedance control, which regulates the stiffness and damping characteristics of the robot’s end-effector to adapt to external forces and achieve compliant motion. However, these methods necessitate explicit modeling of interaction dynamics, making them well-suited for structured environments where system parameters can be accurately defined. As the demand for robotic manipulation in unstructured settings grows, end-to-end deep learning approaches \cite{li2022see, brohan2023rt1roboticstransformerrealworld} have emerged as a promising alternative, offering enhanced adaptability and data-driven force regulation without the need for explicit system modeling. Despite notable advancements, existing research predominantly focuses on improving task success rates, often neglecting the critical aspect of precise force control.

To address this gap, we propose a novel deep learning-based state planning approach to enhance force safety in robotic manipulation. Unlike traditional force control methods, our approach employs pure position control, significantly reducing hardware requirements while maintaining adaptability across diverse unstructured manipulation tasks. From the perspective of state planning, force safety issues mainly arise when the generated state fails to meet the specific physical properties of the structured environments. Taking the door-opening task shown in Fig.~\ref{fig:intro} as an example, the door can only move along the arc-shaped trajectory determined by its physical properties (e.g. the door's size, opening angle, and position relative to the robot). This indicates that all states of the robot’s end-effector, generated by the state planning model, must strictly adhere to this arc-shaped trajectory to ensure force safety. Otherwise, the robot controller will attempt to reach states outside this trajectory, resulting in \textbf{harmful forces}. To be brief, we define the state that lies on this arc-shaped trajectory as \textbf{safe state}, while those outside are deemed unsafe. In this paper, our primary focus is on planning safe states to ensure force safety throughout the robotic manipulation process.

An intuitive solution for the above safe door-opening state planning can be found in bionics: when opening the door, humans estimate future door-opening states based on the door’s physical properties—such as size, opening angle, and so forth—using visual perception, and then modify them in real-time based on the forces sensed through tactile feedback during the actual door-opening process. Inspired by this, we aim to dynamically integrate real-time tactile feedback to refine the vision-guided generated states. However, this solution remains challenging due to the intricate, nonlinear dynamics between the current force feedback and the refinement of future states. These dynamics are influenced by factors like the robot’s manipulability, positions relative to the door, and other physical considerations in the real world.
To address such an issue, we develop a diffusion-based model named \textbf{SafeDiff} to plan safe states, leveraging the effectiveness of diffusion models in approximating complex distributions. In this work, we utilize offline demonstrations collected from the simulator to learn the aforementioned dynamics of the door opening and embed this knowledge into the state representation. This allows us to perform implicit calibration on vision-guided states online, utilizing real-time tactile feedback obtained during inference. Such a process enables the generated states to progressively satisfy the constraints imposed by the door’s properties, thereby ensuring force safety during the entire door-opening process.

This work makes three key contributions: (1) We propose SafeDiff, a diffusion-based model that integrates real-time tactile feedback to implicitly calibrate vision-guided robot states, achieving robustness against external disturbances and maintaining force safety where prior methods often fail. (2) We demonstrate that \textbf{SafeDiff} achieves superior safe state planning across both simulation and real-world experiments, with strong few-shot sim-to-real transfer that greatly reduces real-world data requirements and minimizes object damage risk. (3) We introduce a novel benchmark for force safety in robotic manipulation, including three physically grounded, computationally efficient metrics and the large-scale simulation dataset \textbf{SafeDoorManip50k} for door-opening tasks.

\section{Related Works}
\label{sec:related works}

\subsection{Vision-based Robotic Manipulation}
Numerous studies on vision-based robotic manipulation have addressed tasks such as object grasping~\cite{ichnowski2021dex, dai2023graspnerf}, articulated object manipulation~\cite{mo2019partnet, geng2023rlafford}, and object reorientation~\cite{andrychowicz2020learning}. These works emphasize improving the robot’s environmental perception through various visual modalities to enhance task success rates. For instance, \cite{ichnowski2021dex, dai2023graspnerf, an2024rgbmanip} proposed using RGB-only images for robust robotic manipulation, while SAGCI~\cite{lv2022sagci}, RLAfford~\cite{geng2023rlafford}, and Flowbot3D~\cite{eisner2022flowbot3d} rely solely on point clouds for observations. Additionally, \cite{xu2022universal, wu2020grasp} integrated both RGB images and point clouds to promote the performance on specific manipulation tasks. However, the objects manipulated by robots are often fragile, especially articulated ones. In view of this, vision-based manipulation is challenging to apply in real-world applications because it cannot accurately reflect the force safety status of the manipulated objects. Therefore, it is of great significance for robots to incorporate tactile feedback such that it can dynamically adjust the planned states and handle objects in a safer manner.

\subsection{Multimodal Tactile Feedback for Enhanced Manipulation}
Various learning-based approaches have employed tactile feedback to enhance robotic manipulation. For instance, \cite{murali2018learning} introduced a tactile perception-driven method that enables robots to learn how to grasp objects without relying on visual input. Numerous studies focus on grasp stability \cite{dang2012learning, cui2020self, bekiroglu2013probabilistic}, as well as regrasping~\cite{calandra2018more, su2015force}. A few methods~\cite{van2016stable, kalakrishnan2011learning, sung2017learning, van2015learning} combine reinforcement learning with tactile feedback to formulate manipulation strategies. And very few approaches leverage the combined benefits of both vision and touch. For example, \cite{fu2016one} integrated prior knowledge with dynamic model adaptation to locally compensate for changing dynamics, while~\cite{lee2019making} developed a self-supervised learning framework that fuses visual and tactile inputs for peg insertion, improving learning efficiency. However, the majority of these works used tactile feedback to improve manipulation effectiveness rather than to guide safe planning.

\subsection{Datasets for Door Opening}
In recent years, a primary approach for door manipulation tasks has been to build simulation environments that emulate real-world conditions. Studies such as~\cite{eisner2022flowbot3d, geng2023partmanip, geng2023gapartnet, mo2021where2act, wu2021vat, xu2022universal, zhu2020robosuite} have introduced a variety of simulated door-opening mechanisms, including pushing, pulling, and even those involving latching mechanisms. 
Moreover, datasets like PartNet-Mobility~\cite{xiang2020sapien} and AKB-48~\cite{liu2022akb} offer diverse collections of articulated objects, including doors, but their focus on visual data collection overlooks crucial modalities such as tactile information, limiting their effectiveness for safe door-opening states planning. To address these shortcomings, we developed a comprehensive door manipulation environment with multi-modal inputs and provided a large-scale door-opening dataset to support safe manipulation planning.

\section{Methodology}
\label{sec:method}

\subsection{Preliminary}
We begin by briefly reviewing the diffusion models, a class of generative models that synthesize data by reversing a Markovian process where Gaussian noise is progressively added to data samples. These models consist of two primary phases: the forward process and the reverse process. In the forward process, the original data is systematically corrupted, transitioning from a structured state to pure Gaussian noise over a predefined number of steps, described by the equation $x_t = \sqrt{\alpha_t} x_{t-1} + \sqrt{1-\alpha_t} \epsilon$, where $\epsilon$ is Gaussian noise and $\alpha_t$ are variance-preserving coefficients. The reverse process entails learning to undo the noise addition to recover the original data from its noisy state. This involves training a neural network to estimate the reverse conditional distribution  $p(x_{t-1} | x_t)$, utilizing advanced deep learning techniques.
\begin{figure*}[ht!]
    \centering
    \setlength{\fboxrule}{0pt}
    \framebox{{\includegraphics[width=\linewidth]{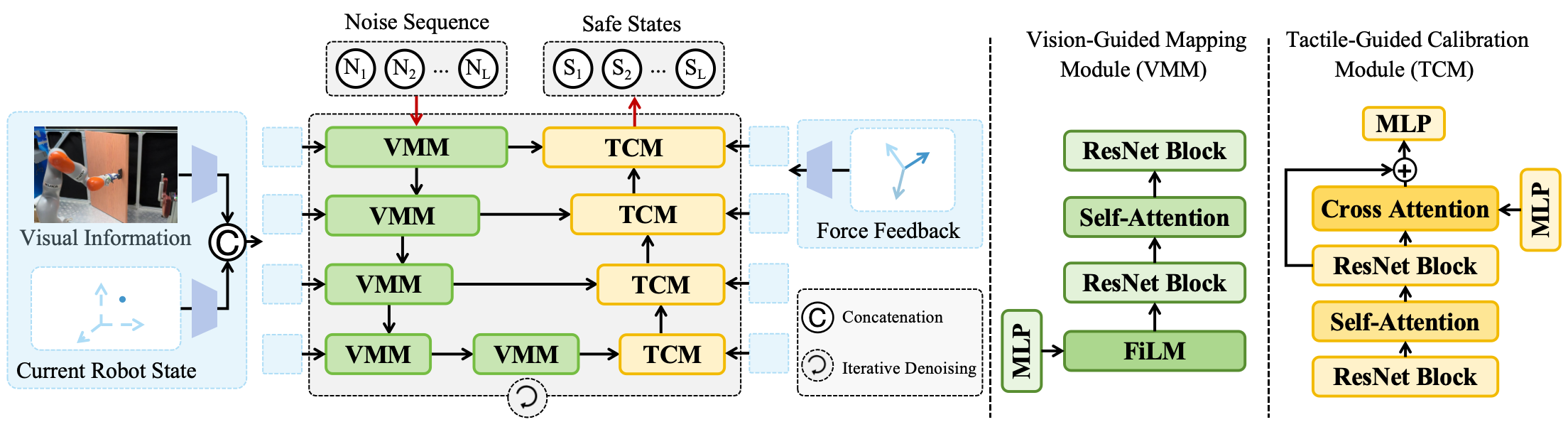}}}
    \caption{Our framework takes a noise sequence as input, visual information, current robot state, and its corresponding force feedback as conditions and outputs the final safe states through $T$ denoising iterations. The architecture consists of an encoder and a decoder. The encoder is composed of a series of multi-scale Vision-Guided Mapping Modules (VMMs) that integrate visual data using FiLM~\cite{perez2018film} and generate state representations initially. The decoder comprises a stack of Tactile-Guided Calibration Modules (TCMs) which can refine the state representations based on tactile feedback.}
    \label{fig:framework}
\end{figure*}
A typical application of the diffusion model in robotic manipulation is Decision	Diffuser~\cite{ajay2022conditional}, which makes decisions using a return-conditional diffusion model, allowing policies to generate behaviors satisfying constraints.

\subsection{The Overall Framework}
Motivated by the Decision	Diffuser~\cite{ajay2022conditional}, the proposed \textbf{SafeDiff} aims to generate a consistent robot state sequence $\mathbf{S}=\{S_k\}_{k=1}^L$ that ensures force safety conditioned on visual-tactile information experienced during manipulation,
thereby preventing any potential damage to the door.
As shown in Fig.~\ref{fig:framework}, we employ an encoder-decoder architecture for our diffusion model. Given the visual representation $\bf{O}$ of the current scene context, typically obtained from the image $\mathbf{I} \in \mathbb{R}^{H \times W \times 3}$, and the current robot state $\mathbf{R} \in \mathbb{R}^{7+6+7+7}$ (representing end effector pose, velocity, and joint position, velocity respectively), the initial input to the model is a set of Gaussian noise ${\mathbf{N} \in \mathbb{R}^{L \times 7}}$ with length $L$. 
After $T$ iterations, the model produces a sequence of $L$ consecutive robot states $\mathbf{S} \in \mathbb{R}^{L \times 7}$. Notably, we have opted to replace the action sequence generated in~\cite{ajay2022conditional} with a sequence of robot states. 
This option stems from the fact that, while the door opening trajectory is predictable, conventional control actions do not inherently guarantee force safety. Instead, each robot state is closely correlated with the current state’s potential harmful force magnitude. Consequently, using robot states facilitates a more robust and efficient model training when integrating tactile feedback.

To harness visual and tactile information effectively to generate safe and reasonable robot states, 
we first introduce the Vision-Guided Mapping Module (VMM) to construct the encoder for our state diffusion model. 
This module translates the robot's current state, denoted as $\hat{\bf{S}}$, and the visual scene context $\bf{O}$, including the door size and relative position to the robot, into a comprehensive state space representation. Although the diffusion model can initially estimate robot state trajectories based on these visual cues, it falls short of guaranteeing force safety during the manipulation process.
To tackle this, we further introduce a Tactile-Guided Calibration Module (TCM) to act as the decoder of our model. Drawing inspiration from human adaptability in responding to tactile feedback and adjusting actions accordingly, this module is designed to capture the intricate, nonlinear dynamics between the current force feedback, represented by $\bf{F}$, and the projected residuals of future states. For more details about the module design, please refer to Sec.~\ref{sec:architecture}.

\subsection{Network Architecture Design}
\label{sec:architecture}
\textbf{Visual-Guided Mapping Module (VMM)} As shown in Fig.~\ref{fig:framework}, we stack a series of VMMs with different temporal scales to construct the encoder for our state diffusion model. In this module, we initially generate the robot state representation by using visual information and the current robot state. Firstly, we use a Multi-Layer Perceptron (\texttt{MLP}), followed by a Resnet block (\texttt{Res}), to extract current scene context from the input image $\mathbf{I}$ and current state $\mathbf{\hat{S}}$. And following FiLM~\cite{perez2018film}, we regard such extracted current scene context as affine coefficients and map Gaussian noise inputs $\mathbf{N}$ into the initial state representation. Then, a self-attention (\texttt{Sttn}) and a Resnet block (\texttt{Res}) are used to enhance the temporal coherence of these state representations:
\begin{align}
[\alpha,~\beta] &= \texttt{MLP}(\mathbf{I},~~\mathbf{\hat{S}}) \\
\mathbf{S^*} &= \texttt{Res}(\alpha \cdot \mathbf{N} + \beta) \\
\mathbf{S^*} &= \texttt{Res}(\texttt{Sttn}(\mathbf{S^*}))
\end{align}
where $\alpha$ and $\beta$ denote the affine coefficients, and $\mathbf{S^*}$ denotes the state representation with the specific temporal scale in the corresponding VMM.

\textbf{Tactile-Guided Calibration Module (TCM)} Similar to the previous module, we utilize a series of TCMs with different temporal scales to form the decoder for our state diffusion model. In this module, we calibrate the robot state representation $\mathbf{S^*}$ to a safer one by introducing tactile information. Before calibration, we use a combination of two \texttt{Res} and one \texttt{Sttn} to further enhance the temporal coherence of $\mathbf{S^*}$. And then, we extract safety context from the input force feedback $\mathbf{F}$ using a \texttt{MLP}. Essentially, harmful forces and states errors can be regarded as 2 physical forms of force insecurity in different spaces (i.e. the former is in force space and the latter is in state space). Based on this, we use a cross attention block (\texttt{Cttn}) to map such extracted safety context into implicit state residual which can be used to calibrate the initial state trajectory generated by the encoder.
\begin{align}
\mathbf{S^*} &= \texttt{Res}(\texttt{Sttn}(\texttt{Res}(\mathbf{S^*}))) \\
\mathbf{S^*} &= \mathbf{S^*} + \texttt{Cttn}(\mathbf{S^*}, \texttt{MLP}(\mathbf{F})) \label{eq:cttn}
\end{align}

\textbf{Implementation} The proposed network follows a multi-scale architecture inspired by U-Net, enabling hierarchical feature extraction and reconstruction. During the encoding phase, VMM is employed for progressive downsampling, successively reducing the sequence length from L to L/2, L/4, L/8, and L/16, while correspondingly increasing the feature dimensions from 3 to 32, 64, 128, and 256. By extracting hierarchical visual features, the encoder captures multi-scale environmental information essential for accurate trajectory generation. In the decoding phase, TCM is utilized for progressive upsampling, integrating tactile features to refine the generated trajectory and enhance adaptability to environmental constraints. To facilitate effective gradient propagation across different scales, shortcut connections are incorporated between all corresponding encoder and decoder layers. This design enhances optimization stability and preserves fine-grained details.

\section{Dataset}
\label{sec:dataset}

To overcome the gap in datasets, we establish the first dataset for ensuring force safety in door opening manipulation planning, named \textbf{SafeDoorManip50k}.
Drawing on the open-source assets detailed in \cite{li2024unidoormanip}, we constructed a diverse collection of $57$ doors, each featuring unique structural designs and distinct color textures. 
Notably, due to functional limitations of the \textit{Isaac Gym}, x-axis harmful forces are inaccurate with the original door handle. Consequently, we made modifications to the collision mesh of the door handle model, enabling accurate readings of the harmful forces in the x-axis. These doors were then divided into a set of $45$ seen doors and a set of $12$ unseen doors. 

In the \textit{Isaac Gym} simulation environment, we established an assembly of doors and robots, where the type, size, and position of the doors, mechanical properties of hinges, stiffness of robots, as well as the lighting conditions, were randomized via random strategies in each scene. 
The label for the sampled demonstration is derived as follows: the door handle's pose in the world coordinate system is accessed via the simulation engine interface, and upon acquiring this pose, the ground truth for the current door opening angle is established by applying the predefined offset between the robot end-effector's and the door handle's coordinate systems.

We sampled a total of $47,727$ training demonstrations on the seen-door set and labeled them accordingly. For testing, employing random strategies akin to those used during training, we sampled $4,580$ scenarios on the seen-door set and $4,438$ on the unseen-door set.

\section{Benchmark}
\subsection{Evaluation Metrics}
We propose a set of novel evaluation metrics specifically designed to comprehensively assess the model’s performance in safe state planning. These metrics address the shortcomings of existing methods for evaluating safe manipulation, offering a more precise and multifaceted assessment of the model’s capabilities.

\textbf{Success Rate (SuR)} 
Unlike~\cite{li2024unidoormanip}, we focus on manipulation force safety by defining SuR as the fraction of test scenarios in which the model completes the task without exceeding 20 N peak force—a threshold chosen under the assumption that forces above 20 N would mechanically damage both the robot and the manipulated object.

\textbf{Average Harmful Force (AHF) and Maximum Harmful Force (MHF)} 
AHF and MHF is applied to evaluate the force-wise force safety of the state planning model in manipulation tasks. It is calculated as the average and maximum harmful force magnitude $\|\textbf{F}_\texttt{harmful}\|$ applied throughout each test process across all test scenarios respectively.

\textbf{Safety Rate (SaR-95 and SaR-80) } 
Safety Rates are utilized to evaluate the scenario-wise force safety of the state planning model in manipulation tasks. 
It is used to ensure that, most of the harmful force magnitudes during the operation remain relatively low, thereby protecting the robot and objects from being continuously exposed to high interaction forces.
Since force safety of a state planning model only depends on the state generated by itself, rather than other states in that trajectory. Thus, we discretize the trajectories with the states planned along the way, i.e.
\begin{align}
\label{eq:forcesafe}
\|\textbf{F}_\texttt{harmful}\|^{k} \leq \textbf{f}, \quad \forall k \in [1, L]
\end{align}
where $L$ denotes the length of states planned by the model and \textbf{f} denotes the force threshold. 
We evaluate the force safety of our state planning model using two metrics: SaR-95 and SaR-80. A test scenario is considered safe under the SaR-95 criterion if $\geq$95\% of its generated states satisfies Eq.~\ref{eq:forcesafe}; similarly, it meets the SaR-80 criterion if $\geq$80\% of the states satisfy this condition. Denoting by \(\mathbf{Num}_{95\%\texttt{safe}}\) and \(\mathbf{Num}_{80\%\texttt{safe}}\) the numbers of test scenarios meeting the SaR-95 and SaR-80 criteria respectively, and by \(\mathbf{Num}_{\texttt{success}}\) the total number of successfully manipulated scenarios, the final metrics are defined as follows:
\begin{align}
\texttt{SaR-95} = \frac{\mathbf{Num}_{95\%\texttt{safe}}}{\mathbf{Num}_{\texttt{success}}},\;\;
\texttt{SaR-80} = \frac{\mathbf{Num}_{80\%\texttt{safe}}}{\mathbf{Num}_{\texttt{success}}}
\end{align}

\subsection{Simulation Experiments} 
\textbf{Implementation}
Our proposed SafeDiff model is implemented based on the publicly available Decision Diffuser code base~\cite{ajay2022conditional}. The training and testing processes are conducted using an NVIDIA A100 Tensor Core GPU.
We utilize the training demonstrations provided by our SafeDoorManip50k for safe state planning. 
The training configuration is as follows: batch size is $256$, total training epochs are $500$, an initial learning rate of $10^{-4}$ with a decay rate of $0.985$, and the application of an Exponential Moving Average (EMA) with a decay factor of $0.995$. During testing, we evaluate the performance of the safe state planning models under $4,580$ seen-door scenarios and $4,438$ unseen-door scenarios in the simulator.
We compare our method with three representative works: the transformer-based multi-modal regression model \cite{li2022see}, the diffusion-based trajectory generator \cite{li2024unidoormanip}, and the action chunking transformer with tactile feedback \cite{li2025hapticactbridginghumanintuition}. For fairness and practicality, we re-implement the latter without its auditory modality.

\begin{table}[ht!]
    \centering
    \renewcommand{\arraystretch}{1.5}
    \caption{Quantitative evaluation of our method and existing models on the \textbf{simulation} scenarios from our \textbf{SafeDoorManip50k}, highlighting the effectiveness of our method in safe state planning. Ours~(V) denotes our method utilizing only visual data as input, while Ours~(V+T) incorporates both visual data and tactile calibration. \checkmark and \ding{55} indicate whether the door manipulated is seen or unseen.}
    \label{tab:QE-sim}
    \resizebox{\textwidth}{!}{
    \begin{tabular}{c|c|c|c|c|c|c|c|c}
        \hline
        \multirow{2}{*}{} & \multirow{2}{*}{Seen (?)} & \multirow{2}{*}{SuR (\%) $\uparrow$} & \multirow{2}{*}{AHF (N) $\downarrow$} & \multirow{2}{*}{MHF (N) $\downarrow$} & \multicolumn{2}{c|}{Threshold - 5 N} & \multicolumn{2}{c}{Threshold - 10 N} \\ \cline{6-9} 
        & & & & & SaR-95 (\%) $\uparrow$ & SaR-80 (\%) $\uparrow$ & SaR-95 (\%) $\uparrow$ & SaR-80 (\%) $\uparrow$ \\ \hline
        Li et al.~\cite{li2022see} & \checkmark & 69.50          &  7.68 & 19.05   &0.10 & 0.66     & 6.09        & 43.12                    \\ 
        Haptic-ACT \cite{li2025hapticactbridginghumanintuition} & \checkmark & 47.10  & 8.41 & 27.32  & 0.00 & 0.04               & 3.12          & 25.18                   \\ 
        UniDoorManip \cite{li2024unidoormanip} & \checkmark & 49.50  & 9.78  & 22.10  & 0.00 & 0.04              & 0.53          & 7.00               \\ 
        Ours (V)  & \checkmark & 78.89  & 6.31  & 16.86 &0.83 & 7.87               & 22.41          & 57.65                   \\ 
        Ours (V+T)  &  \checkmark  & \textbf{80.07}  & \textbf{5.07}  & \textbf{15.03}  & \textbf{6.10}  & \textbf{25.28}  & \textbf{49.25} & \textbf{78.73}   \\ \hline
        Li et al.~\cite{li2022see} & \ding{55} & 68.09           &  7.51    & 18.80    & 0.00 & 0.66            & 8.82             & 47.55               \\
        Haptic-ACT \cite{li2025hapticactbridginghumanintuition} & \ding{55} & 43.94  & 8.57  & 27.34 & 0.02 & 0.49              & 2.40          & 23.05                   \\ 
        UniDoorManip \cite{li2024unidoormanip} & \ding{55} & 52.70  & 9.47  & 21.65 & 0.00 & 0.00              & 0.75          & 11.82                \\ 
        Ours (V)  & \ding{55} & 51.49  & 13.08   & 22.50    & 0.87  & 2.57        & 8.15           & 21.35                 \\ 
        Ours (V+T)  & \ding{55} & \textbf{81.03}           & \textbf{5.08}  & \textbf{14.59} & \textbf{5.13} & \textbf{24.90}     & \textbf{55.54} & \textbf{79.33} \\ \hline
    \end{tabular}
    }
\end{table}

\textbf{Quantitative Results} 
In order to accommodate the limitation of our real experiment, the robot used in our simulated experiment has a fixed base and is stationary. Therefore, the door is considered successfully opened if its angle only surpasses $30^\circ$. In addition, we establish $2$ levels of force thresholds (i.e. $\textbf{f}=5\text{N}~\text{and}~15\text{N}$) to define SaR-95 and SaR-80 in order to evaluate the force safety performance of such involved states planning models more comprehensively.
Tab.~\ref{tab:QE-sim} presents the quantitative results of the models in both the seen-door and unseen-door scenarios discussed earlier. As shown, our method outperforms the others across nearly all metrics. This demonstrates that our method effectively ensures force safety during the robotic manipulation process and can generalize robustly to unseen scenarios.

\textbf{Q1: How does tactile calibration help safe state planning?} 
As tactile calibration plays an essential role in our method, we conduct an ablation study to validate its importance by removing the force feedback input from our method. In the implementation, we directly bypass all operations associated with Eq.~\ref{eq:cttn} during both the training and inference phases. As demonstrated in Tab.~\ref{tab:QE-sim}, without tactile calibration, although our method still manages to successfully open doors, it fails to ensure force safety. More importantly, the absence of tactile calibration significantly impairs our method’s generalization capabilities, which indicates that vision-based state planning methods are inadequate for modeling the intricate dynamics inherent in robotic manipulation tasks, rendering them incapable of planning robustly in dynamic, unstructured environments.

\textbf{Q2: Does SafeDiff still work under environmental disturbances?}
The goal of the \textbf{disturbance} experiment is to observe whether the state planning methods can counteract the environmental disturbances, preventing their accumulation and ultimately avoiding failure in the robotic manipulation tasks.
In the implementation, we tested the involved models using $4,438$ unseen-door scenarios from our SafeDoorManip50k dataset. And during the door-opening process, we applied a periodic impulsive ($1.5\text{Hz}$) disturbance with a positional deviation of $0.03$ meters. Some sample result is visualized in Fig. \ref{fig:dist} of the appendix section \ref{sec:supp-figs}. The (a) is from Ours (V), which fails to overcome the disturbance, and (c) is from Ours (V+T). The (b), (d), (e) are from \cite{li2022see}, \cite{li2025hapticactbridginghumanintuition}, and \cite{li2024unidoormanip} respectively. Our method with tactile calibration responds effectively to the disturbances, maintaining the harmful forces within a relatively small range, and ultimately succeeding in opening the door.

\subsection{Real-world Experiments}
\begin{wrapfigure}{r}{0.50\textwidth} 
\centering
\begin{subfigure}[b]{0.13\textwidth}
    \centering
    \includegraphics[width=\textwidth]{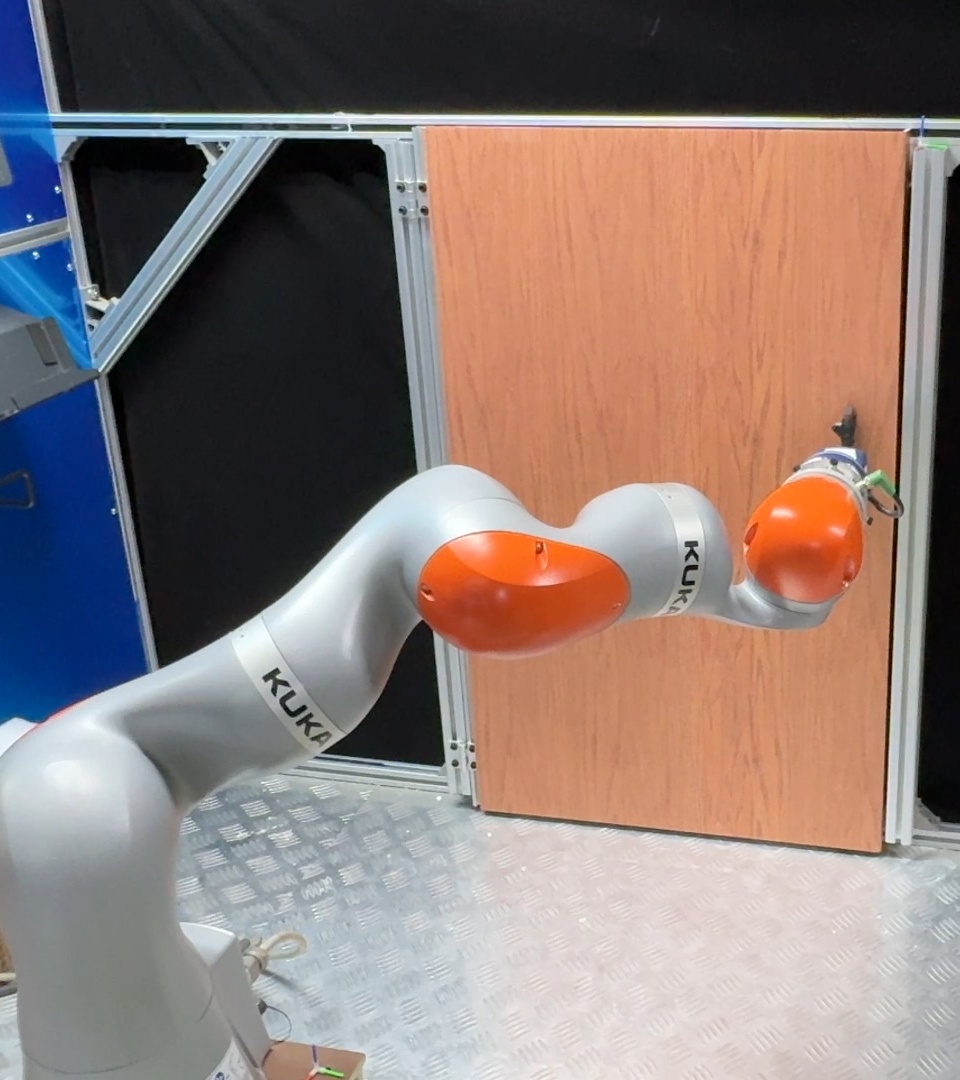}
    \label{fig:realexp1-1}
\end{subfigure}
\hfill
\begin{subfigure}[b]{0.13\textwidth}
    \centering
    \includegraphics[width=\textwidth]{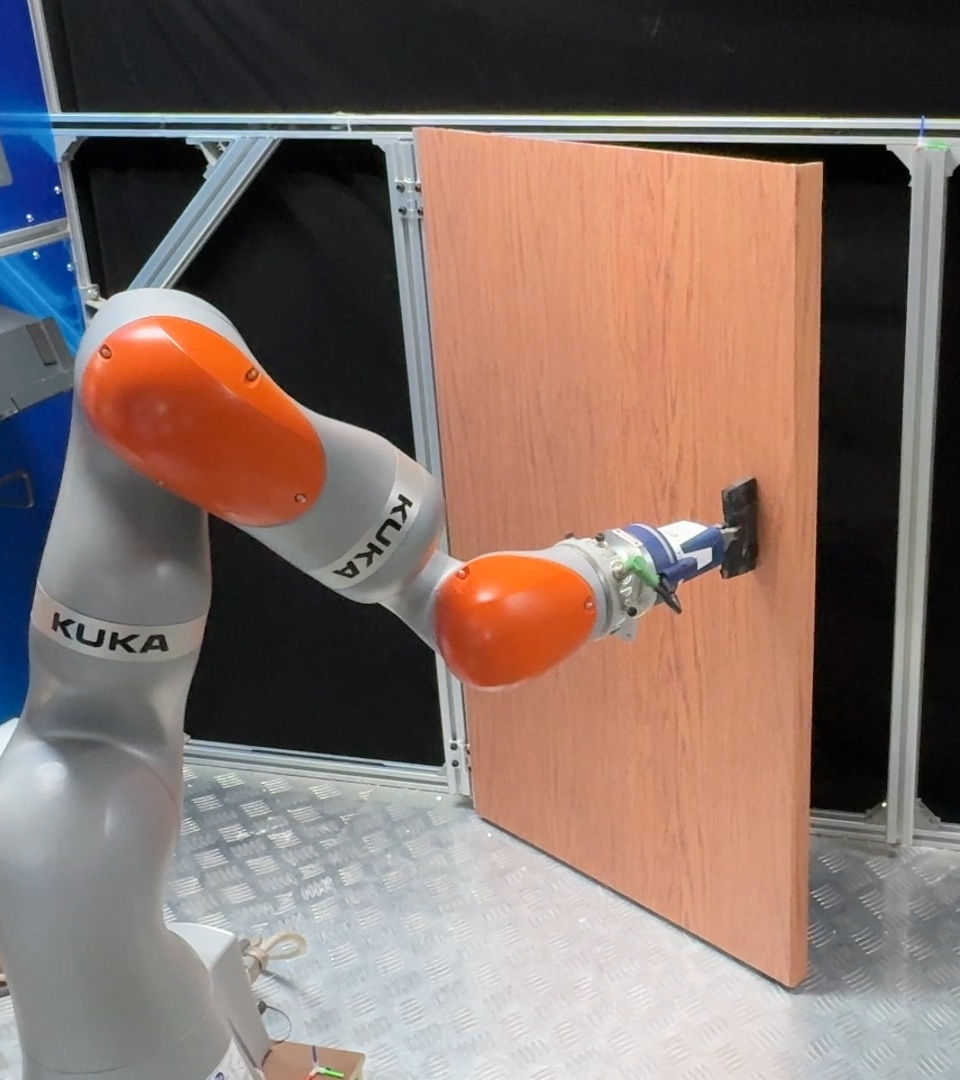}
    \label{fig:realexp1-2}
\end{subfigure}
\hfill
\begin{subfigure}[b]{0.21\textwidth}
    \centering
    \includegraphics[width=\textwidth]{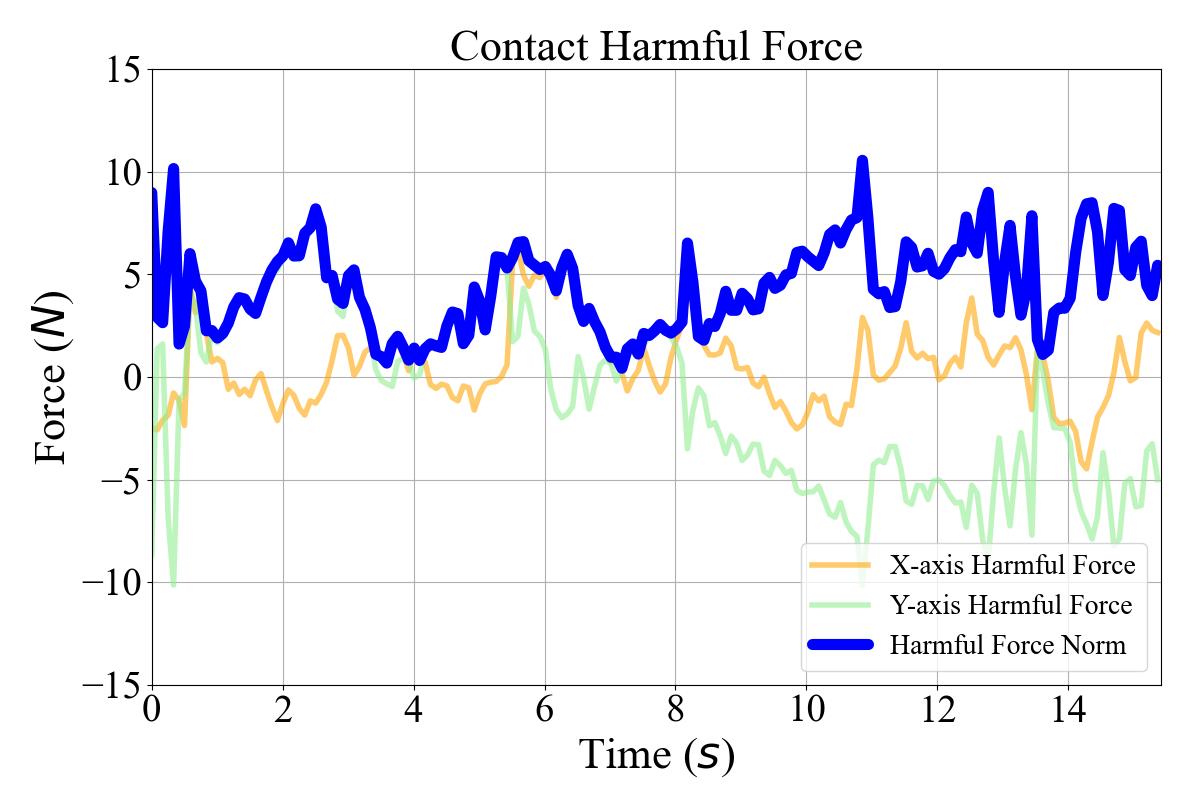}
    \label{fig:realexp1-4}
\end{subfigure}
\vspace{-2mm}

\begin{subfigure}[b]{0.13\textwidth}
    \centering
    \includegraphics[width=\textwidth]{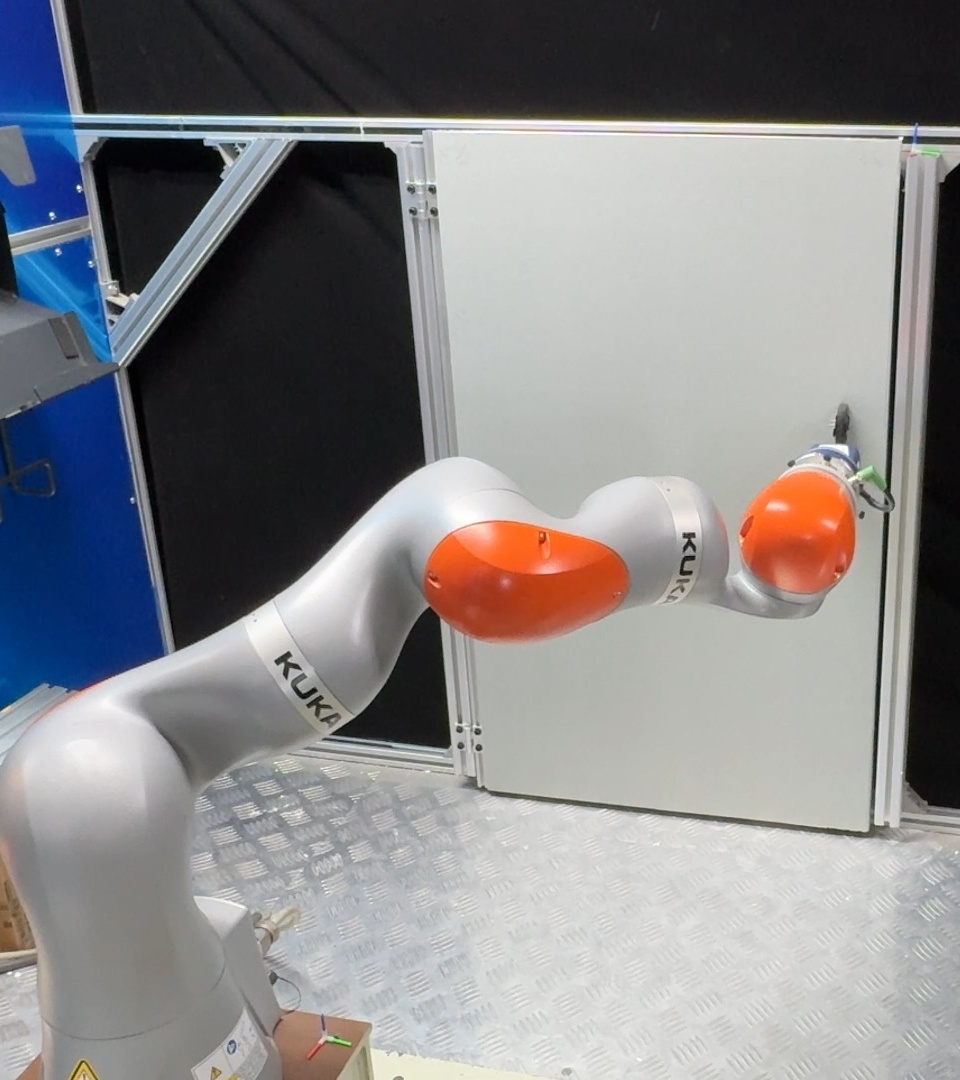}
    \label{fig:realexp2-1}
\end{subfigure}
\hfill
\begin{subfigure}[b]{0.13\textwidth}
    \centering
    \includegraphics[width=\textwidth]{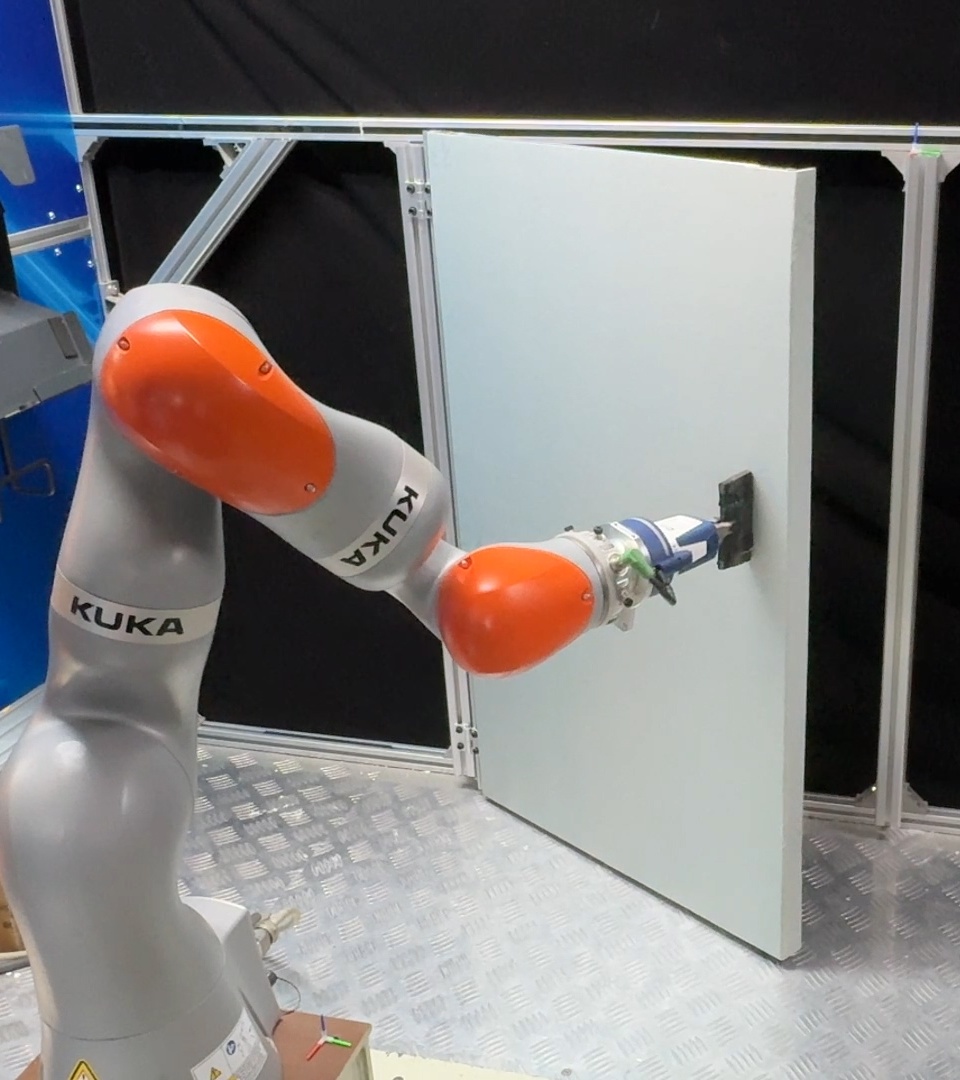}
    \label{fig:realexp2-2}
\end{subfigure}
\hfill
\begin{subfigure}[b]{0.21\textwidth}
    \centering
    \includegraphics[width=\textwidth]{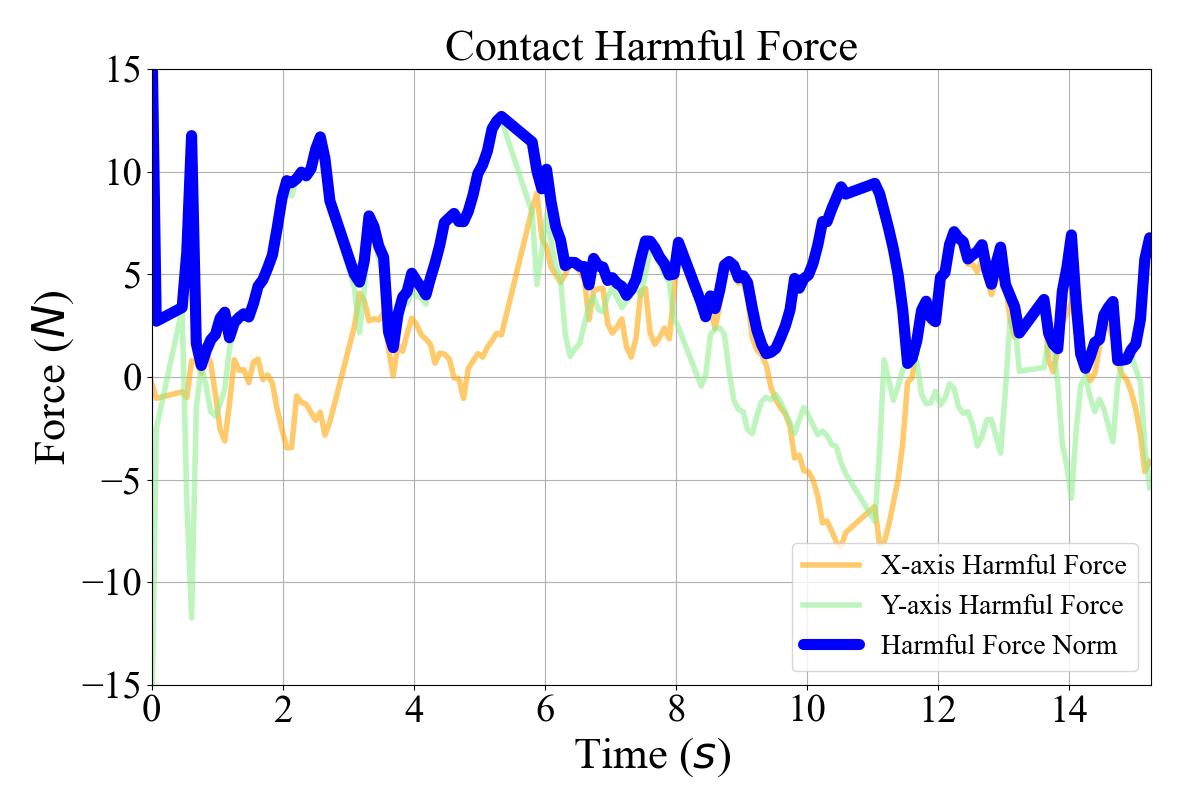}
    \label{fig:realexp2-4}
\end{subfigure}

\vspace{-2mm}

\begin{subfigure}[b]{0.13\textwidth}
    \centering
    \includegraphics[width=\textwidth]{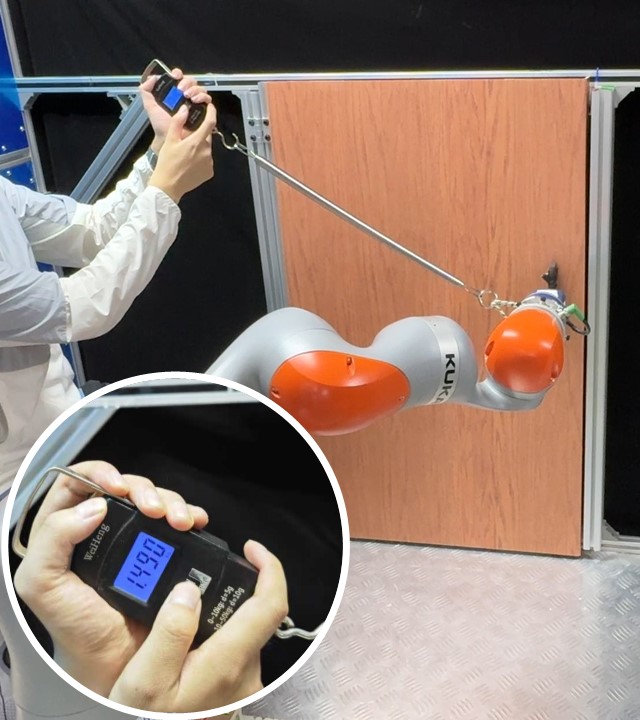}
    \label{fig:realexp3-1}
\end{subfigure}
\hfill
\begin{subfigure}[b]{0.13\textwidth}
    \centering
    \includegraphics[width=\textwidth]{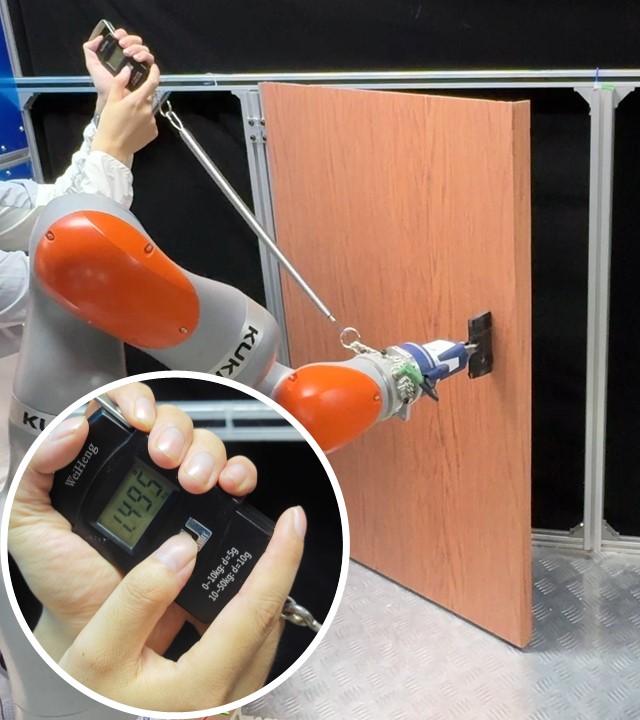}
    \label{fig:realexp3-2}
\end{subfigure}
\hfill
\begin{subfigure}[b]{0.21\textwidth}
    \centering
    \includegraphics[width=\textwidth]{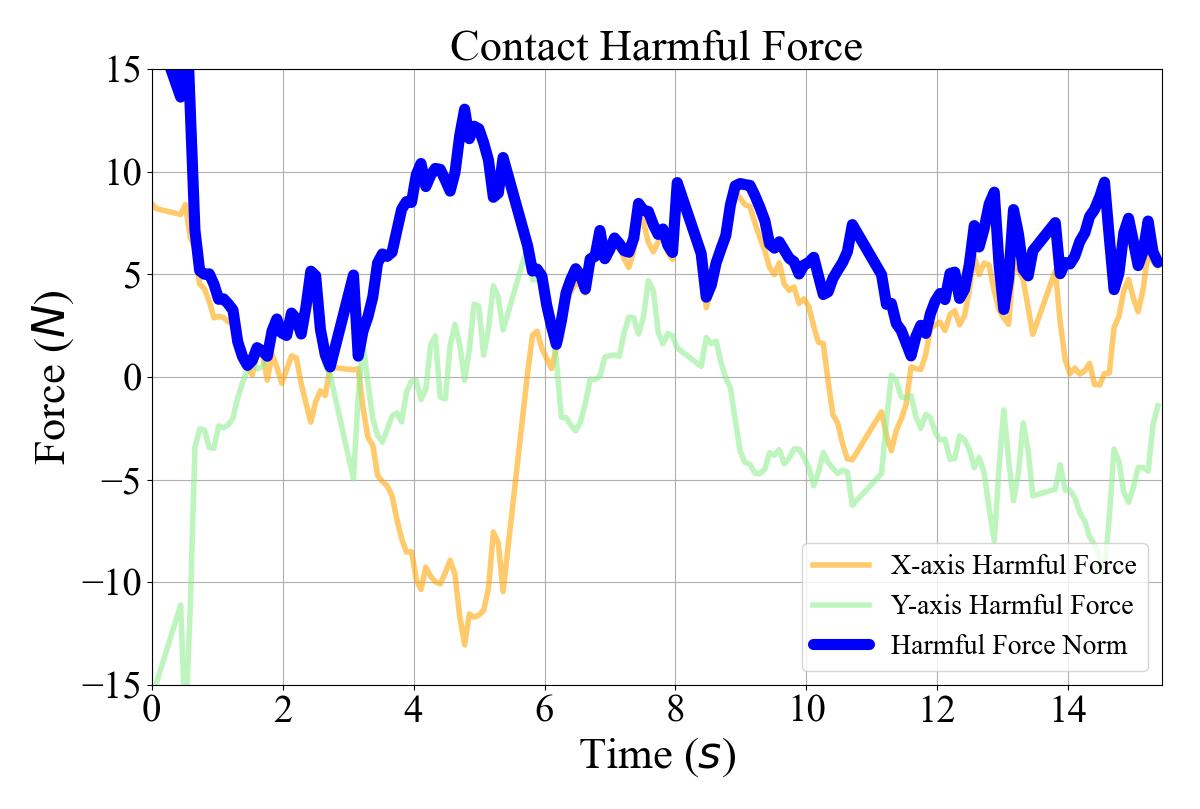}
    \label{fig:realexp3-4}
\end{subfigure}
\caption{Qualitative results of our method in real-world scenarios. Each row corresponds to a specific door-opening task: The first row evaluates the effectiveness of our few-shot fine-tuning model in real-world settings (relevant to \textbf{Q1}), the second row assesses the model’s generalization capabilities (relevant to \textbf{Q2}), and the third row examines the model’s resistance to disturbances (relevant to \textbf{Q3}). Additionally, the first three columns in each row capture two samples from the door-opening process, while the final column quantifies the magnitude of harmful force encountered throughout the entire door-opening. Zoom in $10$ times for the better view.} 
\label{fig:realworldexp}
\end{wrapfigure}

\textbf{Implementation}
In the real-world experiments, we constructed three doors with varying colors and radii. One of these doors was utilized for the collection of training data (referred to as the ``seen" door), while the remaining two were used for unseen tests. Some door samples are shown in Fig.~\ref{fig:realworldexp}. We deployed our state planning model on the KUKA iiwa14 robot. For input of observation, we obtain visual data from an Intel RealSense D435i camera and force feedback from the robot's interior sensors. 
Concurrently, we developed a simulated environment within \textit{Isaac Gym} that closely mirrors the actual environment to gather simulation-augmented data for sim2real experiments. The data collection strategies and labeling methods employed in this experiment were broadly consistent with those used in the simulation.
Ultimately, we collected $110$ real-world demonstrations and $700$ simulation demonstrations.

\textbf{Q1: Can SafeDiff be adapted for real-world robotic manipulation tasks through few-shot fine-tuning?} 
In this experiment, we initially train our model using $700$ sampled simulation demonstrations (denoted as Sim), and subsequently fine-tune it with only $20$ percent of the $110$ real-world demonstrations (denoted as Real~($20\%$)). Fig.~\ref{fig:realworldexp} demonstrates that our method effectively ensures force safety, even with few-shot fine-tuning.

\textbf{Q2: How does the generalization performance of SafeDiff in real-world robotic manipulation tasks through few-shot fine-tuning?}
We continue to employ the few-shot fine-tuned model as the controller for the robot. We then ask the robot to open doors that are unseen during the fine-tuning process. Fig.~\ref{fig:realworldexp} demonstrates that our method exhibits robust generalization capabilities in real-world robotic manipulations.

\textbf{Q3: Does SafeDiff still work under real-world environmental disturbances through few-shot fine-tuning?}
We continue to employ the previously trained model as the robot’s controller. However, unlike in the above experiment, we manually introduce external disturbances during the door-opening process. From Fig.~\ref{fig:realworldexp}, it is evident that our method can effectively calibrate real-world disturbances online, maintaining the harmful force at a low level.

\section{Conclusions}
In this work, we introduce a novel benchmark dedicated to ensuring force safety in robotic manipulation, focusing specifically on manipulation tasks where the robot's motion trajectory is constrained by the physical properties of the manipulated objects, such as door-opening. Drawing inspiration from bionics, we developed a diffusion-based model named \textbf{SafeDiff}, which adeptly integrates real-time tactile feedback to adjust vision-guided planned states, significantly reducing the risk of damage. 
Additionally, we present the \textbf{SafeDoorManip50k} dataset, a pioneering resource that provides a large-scale multimodal environment tailored for safe manipulation. This dataset focuses on the collection of force feedback during robotic manipulation in simulation settings, offering valuable insights that can inspire subsequent tasks. 
Our experiments demonstrate the robust performance of SafeDiff in ensuring safe robotic manipulation.

\noindent\textbf{Limitations.} Given the cost of data collection for simulation and real-world experiments, our experiments are solely conducted on the door-opening task and have not yet been extended to other manipulation tasks. We only consider a gripper rather than a dexterous hand to manipulate objects.
However, we hope that our definition of the evaluation metric, data collection scheme, and model design can stimulate more extensive research in related fields.


\clearpage
\acknowledgments{
This work was partially supported by Shenzhen Science and Technology Program (JCYJ20220818103001002), the Guangdong Key Laboratory of Big Data Analysis and Processing, Sun Yat-sen University, China, and by the High-performance Computing Public Platform (Shenzhen Campus) of Sun Yat-sen University.

The authors would like to express their sincere gratitude to Prof. Yanding Wei for supporting the real-world experiments and for providing valuable feedback on the manuscript.
}


\bibliography{main}  

\newpage
\appendix

\section{SafeDoorManip50k Dataset Details}

\subsection{Door assets production}

\begin{figure}[htbp]
    \centering
    \setlength{\fboxrule}{0pt}
    \framebox{{\includegraphics[width=0.68\textwidth]{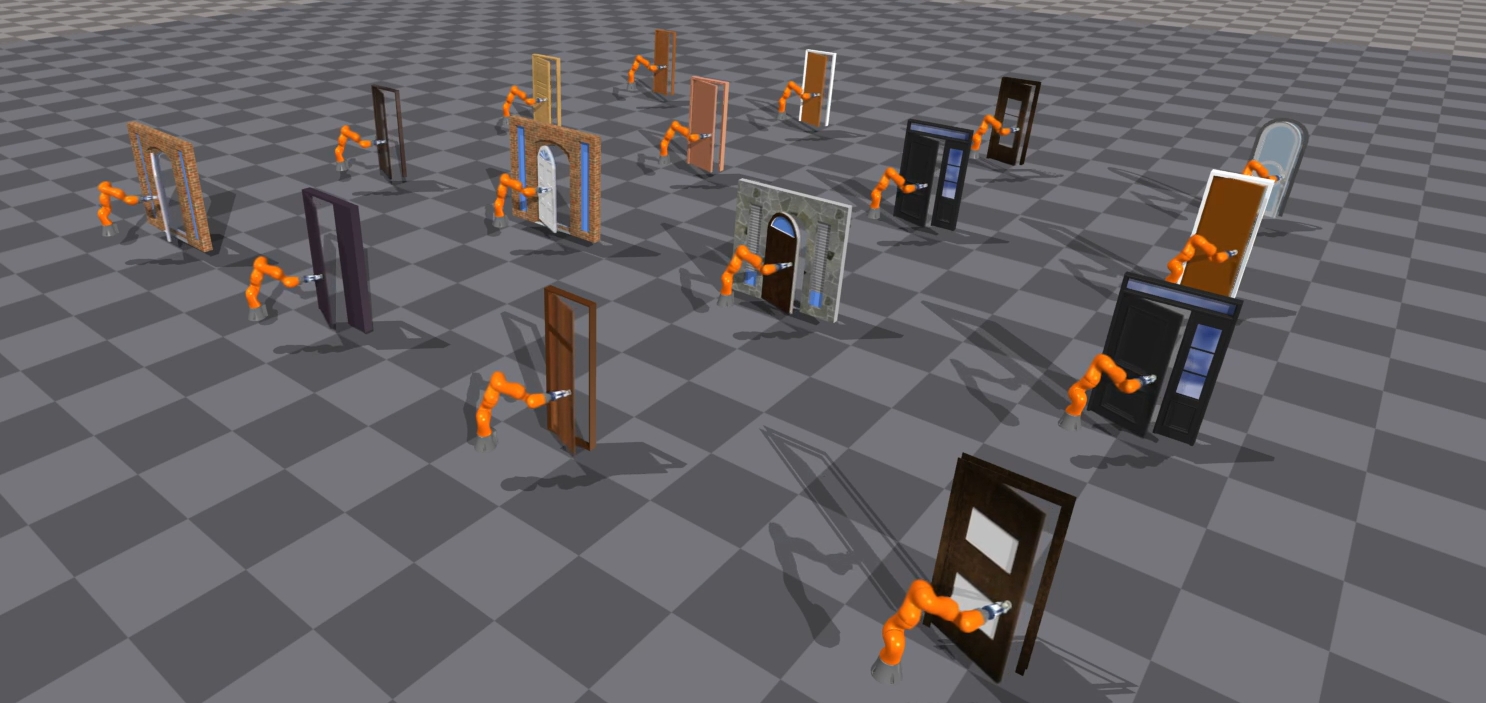}}}
    \caption{Sample of simulation environments}
    \label{fig:app-simenv}
\end{figure}

The door body and handle assets are organized from \cite{li2024unidoormanip}. See Fig. \ref{fig:app-simenv} for some door samples visualization.
We employed the contact force interface within \textit{Isaac Gym} to obtain the contact forces between the robot and the door. Notably, due to functional limitations of the \textit{Isaac Gym} simulation engine, friction force (one source of the contact forces) cannot be read from the \textit{Isaac Gym}'s API. Consequently, we had updated the collision mesh's shape and parameter of the doorknob model, enabling accurate readings of the harmful forces in the horizontal direction. 

\subsection{Data collection configuration}

Leveraging the parallel simulation capabilities of \textit{Isaac Gym}, we simulate 10 distinct door environments per batch. The ground-truth labels for door-pulling trajectories are computed analytically, as the pose of the door handle in the world coordinate system can be directly queried via the \textit{Isaac Gym} API. Given the handle pose, the target label at any specified door opening angle is determined by analytically calculating the handle’s position under the door opening angle with an angular offset.

To facilitate the model’s capability of implicit tactile calibration capabilities, we introduce temporally decaying random positional noise to the target coordinates of the robot’s end effector during execution. The noise direction is uniformly sampled from the interval $[0,\; 2\pi]$, while its magnitude is defined as $A = a\, e^{-k(t - t_0)}$, where $a, e$ are randomized parameters, $t$ denotes the simulation timestep and $t_0$ marks the onset of the current noise perturbation cycle. We perform 5-6 perturbation cycles in each task.

\subsubsection{Random strategies}We provide all random strategy configurations in this part. Here, $\texttt{Uniform}(a,\; b)$ denotes a uniform distribution over the range $[a, b]$, and $\texttt{Gaussian}(\mu,\; \sigma)$ denotes a Gaussian (normal) distribution with mean $\mu$ and standard deviation $\sigma$.

\begin{table}[htbp]
    \centering
    \caption{Parameters and Sampling Distributions in the Simulation Environment}
    \begin{tabular}{@{}ll@{}}
        \toprule
        Parameter & Sampling Distribution \\ \midrule
        Door type & Uniformly sampled from the given door asset set \\
        Door scale & $\texttt{Uniform}(0.8,\; 1.0)$ \\
        Door relative position offset (x, y, z) & $\texttt{Gaussian}(0,\; 0.08)\; m$, $\texttt{Gaussian}(0,\; 0.06)\; m$,\\ & $\texttt{Gaussian}(0,\; 0.06)\; m$ \\
        Door hinge friction & $\texttt{Uniform}(1,\; 10)\; N$ \\
        Door hinge stiffness & $\texttt{Uniform}(1,\; 10)\; N/\mathrm{rad}$ \\
        Robot end-effector stiffness & $\texttt{Gaussian}(3000,\; 100)\; N/m$ \\
        Robot end-effector damping & $\texttt{Gaussian}(100,\; 10)\; N/(m/s)$ \\
        Environment light intensity & $\texttt{Uniform}(0.3,\; 0.7)$ \\
        Perturbation parameter $a$ & $\texttt{Uniform}(3,\; 18) \; mm$\\
        Perturbation parameter $e$ & $\texttt{Uniform}(1.2,\; 1.8)$\\
        \bottomrule
    \end{tabular}
    \label{tab:simulation-parameters}
\end{table}

\section{Real-world Experiment Details}

\subsection{Evaluation Settings}
\label{app:Evaluation Settings in Read-world Experimental Details}
Following the simulation experiment, we establish two levels of force thresholds, $\textbf{F}_\texttt{thres} = 10\text{N}~\text{and}~15\text{N}$, to define SaR-95 and SaR-80. We have omitted $\textbf{F}_\texttt{thres} = 5\text{N}$ due to the real-world noise. In addition, a door is considered successfully opened if its angle exceeds $30^\circ$ while maximum force is smaller than 20N.
Due to limited hardware resources, we were unable to conduct large-scale parallel real-world experiments; however, the results presented demonstrate our model’s effectiveness. During the experiment, three different types of door are used, where one for training and the other two for testing. Each test is repeated 10 times, with other experiment configurations (such as lighting and relative positions between robot and door) are randomized within a certain range. In addition, we chose Li et al.~\cite{li2022see} as the baseline for comparison because it achieved the best performance in simulation.

\begin{table*}[h!]
    \centering
    \renewcommand{\arraystretch}{1.5}
    \caption{Quantitative evaluation of our method and existing models on the \textbf{real-world} scenarios, highlighting the effectiveness of our safe states planning method in the real world. Ours~(V+T) represents our method utilizing both visual data and tactile calibration as inputs. The symbols \checkmark and \ding{55} indicate whether the door manipulated is seen or if there is a disturbance present.}
    \label{tab:QE-real}
    \resizebox{\textwidth}{!}{
    \begin{tabular}{c|c|c|c|c|c|c|c|c|c|c}
        \hline
        \multirow{2}{*}{} 
        & \multirow{2}{*}{Seen~(?)} 
        & \multirow{2}{*}{Disturbance~(?)} 
        & \multirow{2}{*}{Training Set} 
        & \multirow{2}{*}{SuR~(\%)~$\uparrow$} 
        & \multirow{2}{*}{AHF~(N)~$\downarrow$} 
        & \multirow{2}{*}{MHF~(N)~$\downarrow$} 
        & \multicolumn{2}{c|}{Threshold~-~10 N} 
        & \multicolumn{2}{c}{Threshold~-~15 N} \\ \cline{8-11} 
        & & & & & & 
        & SaR-95~(\%)~$\uparrow$ 
        & SaR-80~(\%)~$\uparrow$ 
        & SaR-95~(\%)~$\uparrow$ 
        & SaR-80~(\%)~$\uparrow$ \\ \hline
        Li et al.~\cite{li2022see} 
        & \checkmark & \ding{55} & Real~(100\%) 
        & 100 & 9.038 & 21.781
        & 0 & 0 & 100 & 100 \\
        Ours (V+T) 
        & \checkmark & \ding{55} & Real~(100\%)  
        & 100 & 4.737 & \textbf{16.774}
        & 60 & 100 & 100 & 100 \\
        Ours (V+T) 
        & \checkmark & \ding{55} & Sim~(100\%) + Real~(20\%) 
        & 100 & \textbf{3.763} & 18.581
        & 100 & 100 & 100 & 100 \\ \hline
        Li et al.~\cite{li2022see} 
        & \ding{55} & \ding{55} & Real~(100\%) 
        & 100 & 10.786 & 23.880
        & 0 & 0 & 60 & 90 \\
        Ours (V+T) 
        & \ding{55} & \ding{55} & Real~(100\%)    
        & 100 & 6.564 & 17.339
        & 10 & 50 & 50 & 100 \\ 
        Ours (V+T) 
        & \ding{55} & \ding{55} & Sim~(100\%) + Real~(20\%)  
        & 100 & \textbf{4.709} & \textbf{17.338}
        & 10 & 60 & 60 & 100 \\ \hline
        Li et al.~\cite{li2022see} 
        & \checkmark & \checkmark & Real~(100\%) 
        & 100 & 18.803 & 31.855
        & 0 & 0 & 0 & 0 \\ 
        Ours (V+T) 
        & \checkmark & \checkmark & Real~(100\%)  
        & 100 & \textbf{6.250} & \textbf{28.143}
        & 0 & 100 & 100 & 100 \\ \hline 
    \end{tabular}
    }
\end{table*}

\subsection{Quantitative Results}
\label{app:Quantitative Results in Read-world Experimental Details}
To evaluate the efficacy of the involved models in real-world robotic manipulation, we train them using the entire dataset of 110 real-world demonstrations (denoted as Real~(100\%)).
Each model is then deployed and target points are sent to the robot for the door-opening process. 
The first and second rows of Tab.~\ref{tab:QE-real} show that our method ensures force safety more effectively in real-world robotic manipulation. We elaborate on other aspects in the following discussion.

\textbf{Generalization.}
\label{app:Generalization}
Similar as the simulation experiment, we attempted to open doors that it had not encountered during the fine-tuning stage.
The 2nd and 5th rows of Tab.~\ref{tab:QE-real} demonstrate that our method more effectively in generalization performance. 
Upon further analysis of these experimental results, we observe that the model is more sensitive to changes in door size than to changes in door appearance. Specifically, the robot performs slightly worse on unseen door sizes compared to unseen door faces. However, our implicit tactile calibration successfully corrects the motion trajectory. This observation aligns with the bionic principles discussed in the introduction, further validating the effectiveness of our model design.

\textbf{Anti-disturbance.}
\label{app:Anti-disturbance}
While keeping the previously fine-tuned model unchanged, we manually introduce external disturbances during the door-opening process to assess the real-world disturbance resistance of the models involved. 
Rows 2 and 8 of Table~\ref{tab:QE-real} show that our method can effectively calibrate real-world disturbances online, compared to the baseline method \cite{li2022see} at Row 7.
This indirectly demonstrates that our model adeptly utilizes tactile information as gain-type negative feedback, continuously adjusting the vision-guided planned states. This capability enables robust adaptability to dynamic environmental changes, further validating the effectiveness of our model design.

\textbf{Few-shot Sim-to-Real Transfer.}
\label{app:Few-shot Sim-to-Real Transfer}
To evaluate the efficacy of our model through few-shot fine-tuning, we initially train it using 700 sampled simulation demonstrations (denoted as Sim). We then fine-tune it using only 20\% of the 110 real-world demonstrations (denoted as Real~(20\%)), before deploying on the robot to guide the door-opening process.
Rows 3 and 6 of Table~\ref{tab:QE-real} indicate that our few-shot fine-tuned model outperforms the model trained exclusively on the 110 real-world demonstrations. 
This superior performance can be attributed to the simulated environment’s ability to provide a more intricate and diverse array of training demonstrations, underscoring the importance of our large-scale multimodal simulation dataset, \textbf{SafeDoorManip50k}. This confirms the substantial contribution of such a dataset in enhancing model robustness and adaptability.

\section{Supplementary Figures}
\label{sec:supp-figs}

\begin{figure}[htbp]
\centering
\includegraphics[width=0.9\textwidth]{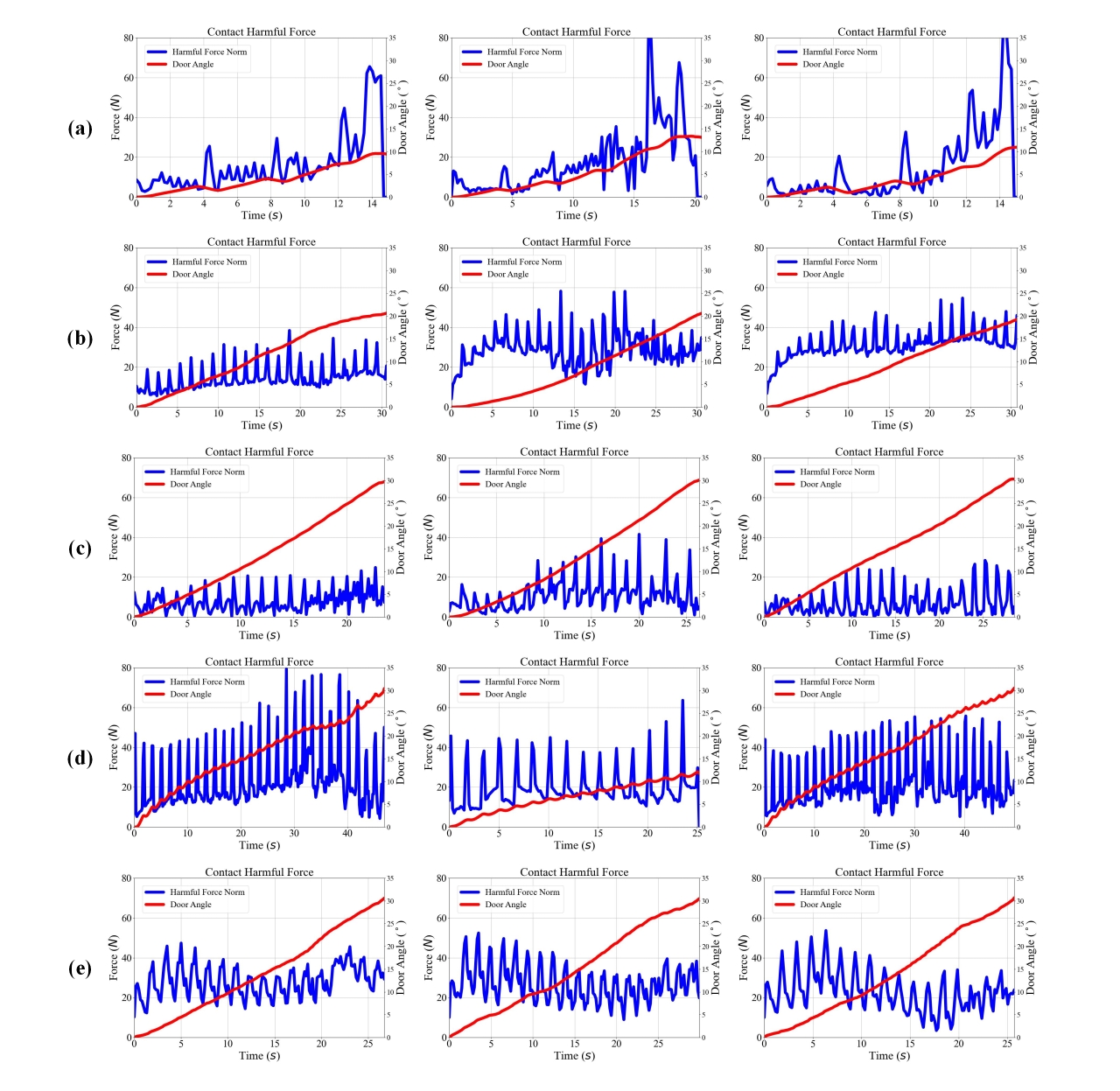} 
\caption{Quantitative evaluation of different methods on our \textbf{SafeDoorManip50k unseen-door} scenarios with \textbf{disturbance}, highlighting the anti-disturbance capability of our method.} 
\label{fig:dist}
\end{figure}



\clearpage




\end{document}